\documentclass[letterpaper, 10 pt, conference]{ieeeconf}  % Comment this line out if you need a4paper
\IEEEoverridecommandlockouts                              % This command is only needed if 
\usepackage{makecell}
\usepackage{cite}
\usepackage{ulem} 
\usepackage{romannum}
\usepackage{balance}
\usepackage{graphicx} %% for loading jpg figures
\usepackage{stfloats}
\usepackage{hyperref}   % to set up hyperlinks
\usepackage{float}
\usepackage[ruled]{algorithm2e}%伪代码用的
\usepackage{caption}
\usepackage{multirow} % Required for multirows
\usepackage{booktabs}%绘制三线表格用的
\usepackage{indentfirst}%首行缩进的
\usepackage{subfigure}
\usepackage{color}  % 改字体颜色的
\usepackage{lipsum} % 输出随机文本
\usepackage{amsmath,esint} % 输出数学公式
\usepackage{cuted}
\usepackage{CJKutf8}
\usepackage{verbatim}
\usepackage{bm}
\usepackage{setspace}
\usepackage{array,caption}
\usepackage{fancyhdr}
\usepackage[labelfont=bf]{caption}
\usepackage{ulem}
\usepackage{color}

\usepackage{arcs}
\usepackage{etoolbox}

\makeatletter
\providecommand\@gobblethree[3]{}
\patchcmd{\over@under@arc}{\@gobbletwo}{\@gobblethree}{}{}
\makeatother

\makeatletter

\newcommand{\Rmnum}[1]{\expandafter\@slowromancap\romannumeral #1@}
\makeatother

\usepackage{CJKutf8}

\hypersetup{
	colorlinks=true,
	linkcolor=blue,
	citecolor=blue,
	urlcolor=blue,
}

\IEEEoverridecommandlockouts                              % This command is only needed if 
\title{\LARGE \bf
Geometric Parameter Optimization of a Novel $3\text{-}(\underset{\bar{}}{\overarc{\text{P}}}\underset{\bar{}}{\text{P}}(2\text{-}(\text{U}\underset{\bar{}}{\text{P}}\text{S})))$ Redundant Parallel Mechanism based on Workspace Determination 
}

\begin{document}
%\begin{CJK}{UTF8}{gkai}
\author{Quan Yuan, Daqian Cao, and Weibang Bai$^{\ast}$
\thanks{
% *This work was not supported by any organization
% This work is supported by the Shanghai Pujiang Program (2023X0203-101-02), the Key Laboratory of Intelligent Perception and Human-Machine Collaboration (ShanghaiTech University), the Ministry of Education,
This work is supported by the Shanghai Pujiang Program under grant 23PJ1408500, by the Shanghai Frontiers Science Center of Human-centered Artificial Intelligence (ShangHAI), MoE Key Laboratory of Intelligent Perception and Human-Machine Collaboration (KLIP-HuMaCo). The experiments of this work were supported by the Core Facility Platform of Computer Science and Communication, SIST, ShanghaiTech University.
% \thanks{$^\dagger$These authors contributed equally to this work.}
% \thanks{$^*$
Corresponding author: Weibang Bai \textit{(wbbai@shanghaitech.edu.cn)}.}
\thanks{Quan Yuan, Daqian Cao, and Weibang Bai are with the ShanghaiTech Automation and Robotics (STAR) Center, School of Information Science and Technology, ShanghaiTech University, Shanghai, 201210, China.}
}

%
% the Startup Funding from ShanghaiTech University}
%\thanks{$^\#$These authors contributed equally to this work. $^{*}$Corresponding author.}
%\textit{(Corresponding author: Weibang Bai)}
% <-this % stops a space

\maketitle
\thispagestyle{empty}
\pagestyle{empty}
% \footnotetext[1]{ShanghaiTech Automation and Robotics (STAR) Center, School of Information Science and Technology, ShanghaiTech University, Shanghai, 201210, China.}
% \renewcommand{\thefootnote}{\fnsymbol{footnote}}
% \footnotetext[1]{These authors contributed equally to this work.}
% \footnotetext[2]{Corresponding author. Email: wbbai@shanghaitech.edu.cn}
% \thispagestyle{plain}  % add page number for the first page
% \pagestyle{plain}      % add page number for the second and the later pages
%%%%%%%%%%%%%%%%%%%%%%%%%%%%%%%%%%%%%%%%%%%%%%%%%%%%%%%%%%%%%%%%%%%%%%%%%%%%%%%%
\begin{abstract}
Redundant parallel robots are normally employed in scenarios requiring good precision, high load capability, and large workspace compared to traditional parallel mechanisms. However, the elementary robotic configuration and geometric parameter optimization are still quite challenging. This paper proposes a novel $3\text{-}(\underset{\bar{}}{\overarc{\text{P}}}\underset{\bar{}}{\text{P}}(2\text{-}(\text{U}\underset{\bar{}}{\text{P}}\text{S})))$ redundant parallel mechanism first,
% with good generalizability first, 
and further investigates the kinematic optimization issue by analyzing and investigating how its key geometric parameters influence 
% \sout{its workspace characteristics including the workspace} 
the volume, shape, boundary completeness, and orientation capabilities of its workspace. The torsional capability index ($\rm TI_1$) and tilting capability index ($\rm TI_2$) are defined to evaluate the orientation performance of the mechanism. Numerical simulation studies are completed to indicate the analysis, providing reasonable but essential references for the parameter optimization of $3\text{-}(\underset{\bar{}}{\overarc{\text{P}}}\underset{\bar{}}{\text{P}}(2\text{-}(\text{U}\underset{\bar{}}{\text{P}}\text{S})))$ and other similar redundant parallel mechanisms.

% effect of the parametric variation, the workspace characteristics change predictably. This provides guidance for the efficient selection of geometric parameters for $6 - \rm (\bar P\bar P)U\bar PU$ and other redundant parallel mechanisms with similar structures.

% \begin{equation}
% \alpha_{20}=\pi-\arcsin \left(\frac{L_{1}}{R_{1}+R_{2}}\right)
% \end{equation}

\end{abstract}

%%%%%%%%%%%%%%%%%%%%%%%%%%%%%%%%%%%%%%%%%%%%%%%%%%%%%%%%%%%%%%%%%%%%%%%%%%%%%%%%
\section{Introduction}

Parallel robots have been widely applied in numerous fields due to their advantages of high load capacity and precision \cite{zhao2020evolution,gosselin2016kinematically,tang2012novel}. However, the limitations of parallel robots, such as small workspaces and complex singularities have restricted their practical applications in specific areas, such as medical surgery and rehabilitation \cite{lacombe2022singularity}. 

Introducing redundancy, including kinematic redundancy and actuation redundancy, into a parallel mechanism is an effective approach to enlarge its workspace and reduce singularity problems \cite{luces2017review, gosselin2018redundancy}. Abadi et al. \cite{abadi2019redundancy} developed a novel spatial parallel mechanism with kinematic redundancy, introducing an online redundancy resolution and control strategy to avoid singularities and enhance dynamic performance through real-time trajectory optimization. Yuan et al. \cite{yuan2024development} developed a novel redundant parallel mechanism with an enlarged workspace and enhanced dexterity for fracture reduction surgery. Tian et al. \cite{tian2021design} proposed a novel kinematically redundant reconfigurable generalized parallel manipulator with configurable platforms, introducing a new design to improve workspace, dexterity, and adaptability for diverse tasks. 

However, the robotic optimization design becomes increasingly challenging due to the introduced redundancy, especially when the compromise between the functional workspace and the structural size of the robot is commonly inquired.
% is becoming more problematic 
% However, robotic optimization design becomes increasingly challenging due to the introduced redundancy, especially when balancing the functional workspace with the structural size of the robot is a key consideration.
% of the parallel robots is normally established with the performance indices and constraint conditions defined. 
% Furthermore, 
With some explicit indices defined, multi-objective optimization can be employed to simultaneously enhance the selected performance for parallel robot design \cite{yang2022review, liang2023kinematic, clancy2024analysis}. 
Nevertheless, the relationship between the workspace representation and the robotic kinematic parameters is general and intuitive for robotic design, but it is implicit which makes the parameter optimization problem difficult to define \cite{zhi2021kinematic, cursi2022globdesopt}, thus problematic to solve using various optimization algorithms.
% like Genetic Algorithms \cite{wu2016architecture}, particle swarm optimization algorithms \cite{wang2017optimal},  etc.

% To optimize the parameters and achieve large dexterous workspace, the basic mapping relationship between the workspace features and the kinematic parameters of the robot should be addressed or estimated. 
%
% Robotic optimization design always needs to establish the objective performance indices and relevant constraints. 

In this field, researchers have presented some relevant explorations. Zhi et al. \cite{zhi2021kinematic} proposed a parameter optimization method for a serial miniaturized surgical instrument only based on dexterous workspace volume estimation. Wu et al. \cite{wu2016architecture} focused on the architecture optimization of a parallel Schönflies-motion robot using a genetic algorithm to enhance performance within a predefined workspace for pick-and-place applications. Wang et al. \cite{wang2017optimal} employed a multi-objective particle swarm optimization algorithm to optimize the workspace and natural frequency of a planar parallel 3-DOF nanopositioner. 
% Su et al. \cite{su2024design} used genetic algorithms to optimize the overall performance of their proposed redundant parallel fracture reduction robot. 

Nonetheless, the aforementioned studies normally take the workspace volume as an important optimization objective, while lacking comprehensive investigations on more concerns of the workspace characteristics like the shape, boundary
completeness, and orientation capabilities. Moreover, lots of the work utilizing optimization algorithms for complex parallel mechanisms only tried to find the optimal design parameters for specific targets, revealing no causal relationship between the core design variables.
%Meanwhile, many studies employing optimization algorithms for complex parallel mechanisms focus solely on identifying optimal design parameters for specific targets, without revealing any intuitive causal relationships for general design.
% between the core design variables.

% based on specific explicit objectives, which are not general and intuitive. 

% especially redundant designs, can be computationally time-consuming, making it difficult to select optimal geometric parameters intuitively.

To this end, this paper focuses on systematically investigating the influence of key geometric parameters on the workspace characteristics of a novel $3\text{-}(\underset{\bar{}}{\overarc{\text{P}}}\underset{\bar{}}{\text{P}}(2\text{-}(\text{U}\underset{\bar{}}{\text{P}}\text{S})))$ redundant parallel mechanism. 
The main contributions can be summarised as:
    \begin{itemize}
        \item A novel $3\text{-}(\underset{\bar{}}{\overarc{\text{P}}}\underset{\bar{}}{\text{P}}(2\text{-}(\text{U}\underset{\bar{}}{\text{P}}\text{S})))$ redundant parallel mechanism is proposed, and its detailed kinematics is analyzed.
        \item An orientation workspace representation method based on the axes of symmetry of the mechanism is introduced and adopted.
        \item The effects of key geometric parameters on the workspace characteristics of the novel $3\text{-}(\underset{\bar{}}{\overarc{\text{P}}}\underset{\bar{}}{\text{P}}(2\text{-}(\text{U}\underset{\bar{}}{\text{P}}\text{S})))$ mechanism are investigated, which provides an intuitive parameter optimization solution for redundant parallel mechanisms based on the workspace determination.
    \end{itemize}

\section{Redundant parallel mechanism $3\text{-}(\underset{\bar{}}{\overarc{\text{P}}}\underset{\bar{}}{\text{P}}(2\text{-}(\text{U}\underset{\bar{}}{\text{P}}\text{S})))$}\label{Mech}
As stated previously, introducing redundancy into parallel mechanisms can enlarge their workspace and benefit the singularity avoidance issue. Consequently, we propose a novel redundant parallel mechanism by incorporating 6 prismatic joints to the fixed base, which can be used in various scenarios like robotic-assisted orthopedic surgery. However, the parameter optimization design of redundant parallel robots remains a fundamental challenging problem. Thus, instead of explaining the origination of its configuration innovation, we mainly focus on its general mechanism analysis and the subsequent parameter optimization in this work.
% Without loss of generality,  are mainly focused on here.

\subsection{Design of Redundant Parallel Mechanism $3\text{-}(\underset{\bar{}}{\overarc{\text{P}}}\underset{\bar{}}{\text{P}}(2\text{-}(\text{U}\underset{\bar{}}{\text{P}}\text{S})))$}

\begin{figure}[h]
  \centering
    \includegraphics[width=3.3in]{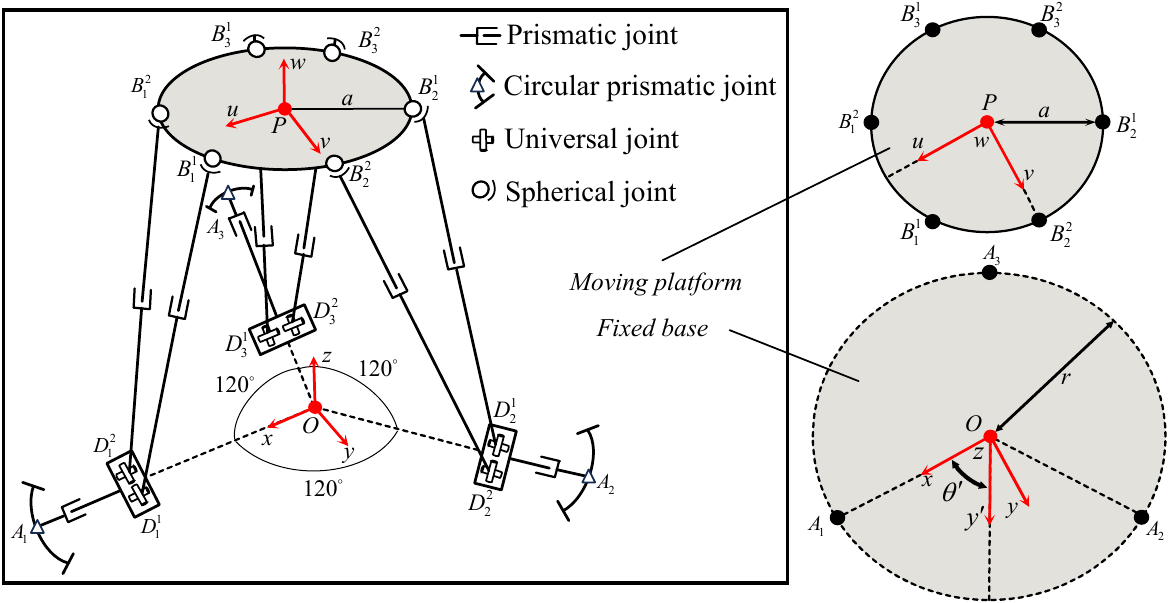}
    \caption{Kinematic diagram of $3\text{-}(\underset{\bar{}}{\overarc{\text{P}}}\underset{\bar{}}{\text{P}}(2\text{-}(\text{U}\underset{\bar{}}{\text{P}}\text{S})))$.}\label{fig1}
\end{figure}

Firstly, a novel redundant parallel mechanism incorporating 3 circular prismatic joints and 3 prismatic joints to the fixed base of traditional parallel robot is proposed, which is $3\text{-}(\underset{\bar{}}{\overarc{\text{P}}}\underset{\bar{}}{\text{P}}(2\text{-}(\text{U}\underset{\bar{}}{\text{P}}\text{S})))$, as shown in Fig. \ref{fig1}. 
The coordinate $O-xyz$ is established at the center point $O$ of the fixed base, where the $x$-axis points to ${A_1}$ and the $z$-axis is perpendicular to the fixed base. The coordinate $P-uvw$ is established at the center point $P$ of the moving platform, where ${B_1^2}$ and ${B_1^1}$ are symmetric with respect to the $u$-axis, the $v$-axis points to ${B_2^2}$, and the $w$-axis is perpendicular to the moving platform. ${B_i^j(i=1,2;j=1,2,3)}$ and ${A_i(i=1,2,3)}$ are evenly set on the moving platform and fixed base, respectively. The radius of the moving platform is $a$. The distance between ${D_1^2}$ and ${D_1^1}$ is $b$. $\rm \bar P$ represents the active prismatic joint.

The mobility $M$ of $3\text{-}(\underset{\bar{}}{\overarc{\text{P}}}\underset{\bar{}}{\text{P}}(2\text{-}(\text{U}\underset{\bar{}}{\text{P}}\text{S})))$ can be calculated according to the modified Grübler-Kutzbach criterion\cite{li2013applicability}:
\begin{equation}
% \begin{align}
M = d(n - g - 1) + \sum\limits_{k = 1}^g {{f_k} + } \upsilon  - \xi
% \end{align}
\end{equation}
where the value of $d$ for the spatial mechanism is six, which represents the order of the mechanism. $n$ is the number of components, $g$ is the number of kinematic pairs, $f_i$ is the degree of freedom (DOF) of the $i$-th kinematic pair, $\upsilon $ is the number of the excessive non-common constraints, and $\xi $ is the number of passive DOFs in this mechanism. Thus, we have $M=6(20 - 24 - 1) + 42 + 0 - 0 = 12$.

\subsection{Inverse Kinematic Analysis}
To find the inverse kinematics (IK) solution for the moving platform, the orientation of the moving platform in terms of Euler angles in the z-y-x sequence can be represented as ($\psi$, $\theta$, $\varphi$). The transformation matrix that relates the moving platform to the fixed base can be written as
\begin{equation}
	\begin{array}{l}
		% {\kern 1pt} {\kern 1pt} {\kern 1pt} {\kern 1pt} {\kern 1pt} {\kern 1pt} {\kern 1pt} {\kern 1pt} {\kern 1pt} {\kern 1pt} {\kern 1pt} {\kern 1pt} {\kern 1pt} {\kern 1pt} {\kern 1pt} {\kern 1pt} {\kern 1pt} {\kern 1pt} {\kern 1pt} {\kern 1pt} {\kern 1pt} {\kern 1pt} {\kern 1pt} {\kern 1pt} {\kern 1pt} {\kern 1pt} {\kern 1pt} {\kern 1pt} {\kern 1pt} {\kern 1pt} {\kern 1pt} {\kern 1pt} {\kern 1pt} {\kern 1pt} {\kern 1pt} {\kern 1pt} {\kern 1pt} {\kern 1pt} {\kern 1pt} {\kern 1pt} {\kern 1pt} {\kern 1pt} {\kern 1pt} {\kern 1pt} {\kern 1pt} {\kern 1pt} {\kern 1pt} {\kern 1pt} {\kern 1pt} {\kern 1pt} {\kern 1pt} {\kern 1pt} {\kern 1pt} 
  {{\bf{R}}_1} = {{\bf{R}}_z}(\psi ){{\bf{R}}_y}(\theta ){{\bf{R}}_x}(\varphi ) = \\
		\left[ {\begin{array}{*{20}{c}}
				{c\psi c\theta }&{\quad  - s\psi c\varphi  + c\psi s\theta s\varphi }&{\quad s\psi s\varphi  + c\psi s\theta c\varphi }\\
				{s\psi c\theta }&{\quad c\psi c\varphi  + s\psi s\theta s\varphi }&{\quad  - c\psi s\varphi  + s\psi s\theta c\varphi }\\
				{ - s\theta }&{\quad c\theta s\varphi }&{\quad c\theta c\varphi }
		\end{array}} \right]
	\end{array}
\end{equation}
where $s$ and $c$ represent sine and cosine respectively.

The position vector of $B_i^j(i = 1,2,3;j = 1,2)$ in the fixed base can be expressed as
\begin{equation}
    {\bm B}_i^j = {{\bm P}_o} + { {\bm R}_1}\bar  {\bm u}_i^j \label{eq:3}
\end{equation}
%
%\redcolor{Note: seems ${}^1{\bm B}_i^j$ is a point, not a vector? or with respect to which point in this case?}
%
%\redcolor{Reply: In this manuscript, bold letters represent vectors, while non-bold letters denote scalars or point names. For example, ${}^1B_i^j$ refers to a point, and ${}^1{\bm B}_i^j$ denotes the position vector of ${}^1B_i^j$ with respect to the fixed base, such as the vector ${\bm O}{}^1{\bm B}_i^j$}.
%
where ${{\bm P}_o}$ is the position vector of ${P}$ in the fixed base. $\bar {\bm u}_i^j$ is the position vector of ${{ B}_i^j}$ in the moving platform. It can be expressed as
\begin{equation}
\bar {\bm u}_i^j = {{\bm Z}_i}{\left[ {\begin{array}{*{20}{c}}
{\frac{{\sqrt 3 a}}{2}}&{\frac{{{{( - 1)}^{j + 1}}a}}{2}}&0
\end{array}} \right]^{\rm T}}
\end{equation}
where
\begin{equation}
    {{\bm Z}_i} = \left[ {\begin{array}{*{20}{c}}
{c{\gamma _i}}&{ - s{\gamma _i}}&0\\
{s{\gamma _i}}&{c{\gamma _i}}&0\\
0&0&1
\end{array}} \right] \label{eq:5}
\end{equation}
where $\gamma_i=(i-1)2\pi/3$ is the angle between $OA_i$ and $x$-axis in the fixed base.

The position vector of $B_i^j$ in the fixed base can be written as
\begin{equation}
     {{\bm u}_i^j}={{\bm R}_1}\bar {\bm u}_i^j
\end{equation}

The relationship between the angular velocity and the derivative of the orientation of the moving platform in the fixed base can be expressed as 
\begin{equation}
    {\bm \omega}  = {\bm{R}_2} \bm {\dot \Theta}
\end{equation}
where $ \bm {\dot \Theta}  = {\left[ {\begin{array}{*{20}{c}}
\varphi &\theta &\psi 
\end{array}} \right]^ {\rm T}}$ denotes the orientation of the moving platform, and $\bm R_2$ can be written as
\begin{equation}
    {{\bm R}_2} = \left[ {\begin{array}{*{20}{c}}
{c\theta c\varphi }&{ - s\varphi }&0\\
{c\theta s\varphi }&{c\varphi }&0\\
{ - s\theta }&0&1
\end{array}} \right]
\end{equation}

The velocity of $ B_i^j$ can be calculated by the time derivative of Eq. (\ref{eq:3})
\begin{equation}
\dot {\bm B}_i^j = {\dot {\bm P}_o} + \bm \omega  \times \bar {\bm u}_i^j \label{eq:9}
\end{equation}
Eq. (\ref{eq:9}) can be rewritten as
\begin{equation}
\dot {\bm B}_i^j = \left[ {\begin{array}{*{20}{c}}
{{{\bm I}_{3 \times 3}}}&{ - \tilde {\bm u}_i^j{{\bm R}_2}}
\end{array}} \right]\left[ {\begin{array}{*{20}{c}}
{{{\dot {\bm P}}_o}}\\
{ \dot {\bm \Theta} }
\end{array}} \right] \label{eq:10}
\end{equation}
where $\tilde {\bm u}_i^j$ is the skew-symmetric matrix of ${\bm u}_i^j$.

The position vector of $D_i^j$ in the fixed base can be expressed as
\begin{equation}
{\bm D}_i^j = {{\bm Z'}_i}{\left[ {\begin{array}{*{20}{c}}
{r - d_i}&{\frac{{{{( - 1)}^{j + 1}}b}}{2}}&{0}
\end{array}} \right]^{\rm T}} \label{eq:11}
\end{equation}
where
\begin{equation}
{{\bm Z'}_i} = \left[ {\begin{array}{*{20}{c}}
{c\varpi_i }&{ - s\varpi_i}&0\\
{s\varpi_i}&{c\varpi_i}&0\\
0&0&1
\end{array}} \right] 
\end{equation}
where $d_i$ denotes the variable of the prismatic joint on the fixed base, $r$ denotes the radius of the fixed base, $\eta_i$ denotes the rotation angle of the circular prismatic joint, and $\varpi_i=\eta_i+\gamma_i$.

The time derivative of Eq. (\ref{eq:11}) can be derived as 
\begin{equation}
    \dot {\bm D}_i^j = {{\bm h}_i}\dot {d}_i + {\bm t}_i\dot \eta _i \label{eq:12}
\end{equation}
where
\begin{equation}
{{\bm h}_i} = {\left[ { - \begin{array}{*{20}{c}}
{{\mathop{\rm c}\nolimits} {\varpi _i}}&{ - s{\varpi _i}}&0
\end{array}} \right]^{\rm T}}
\end{equation}
and
\begin{equation}
{{\bm t}_i} = \left[ {\begin{array}{*{20}{c}}
{({d_i} - r)s{\varpi _i} - \frac{{{{( - 1)}^{j + 1}}b}}{2}c{\varpi _i}}\\
{(r - {d_i})s{\varpi _i} - \frac{{{{( - 1)}^{j + 1}}b}}{2}s{\varpi _i}}\\
0
\end{array}} \right]
\end{equation}

Based on Eq. (\ref{eq:3}) and Eq. (\ref{eq:11}), the length of the prismatic joint between $B_i^j$ and $D_i^j$ can be calculated by
\begin{equation}
    l_i^j = \left\| {{\bm B}_i^j - {\bm D}_i^j} \right\|
\end{equation}

${\bm B}_i^j$ can be rewritten as
\begin{equation}
    {\bm B}_i^j = {\bm D}_i^j + l_i^j{\bm s}_i^j \label{eq:16}
\end{equation}
where ${\bm s}_i^j$ denotes the unit vector of $\bm B_i^j \bm D_i^j$.

The time derivative of Eq. (\ref{eq:16}) can be derived as
\begin{equation}
    \left[ {\begin{array}{*{20}{c}}
{{{\bm I}_{3 \times 3}}}&{ - \tilde {\bm u}_i^j{{\bm R}_2}}
\end{array}} \right]\left[ {\begin{array}{*{20}{c}}
{{{\dot {\bm P}}_o}}\\
{ \dot {\bm \Theta} }
\end{array}} \right] = {\bm {h}_i}\dot d_i + {\bm t}_i{}\dot \eta _i + {\bm s}_i^j \dot l_i^j +  {\bm S}_i^j \times {\bm s}_i^j l_i^j\label{eq:17}
\end{equation}
where ${\bm S}_i^j$ denotes the angular velocity of $\bm B_i^j \bm D_i^j$.

Then multiply ${\bm s}_i^j$ on both sides of Eq. (\ref{eq:17}), and since $\bm s_i^j\cdot(\bm S_i^j\times \bm s_i^j)=0$, $\dot l_i^j$ can be expressed as
\begin{equation}
\dot l_i^j = {\bm s}_i^{j\rm T}\left[ {\begin{array}{*{20}{c}}
{{{\bm I}_{3 \times 3}}}&{ - \tilde {\bm u}_i^j{{\bm R}_2}}
\end{array}} \right]\left[ {\begin{array}{*{20}{c}}
{{{\dot {\bm P}}_o}}\\
{ \dot {\bm \Theta} }
\end{array}} \right] - {\bm s}_i^{j\rm T}{{\bm h}_i}\dot d_i - {\bm s}_i^{j\rm T}{\bm t}_i\dot \eta _i \label{eq:18}
\end{equation}

The position and orientation of the moving platform are denoted by
\begin{equation}
    {\bm x} = {\left[ {\begin{array}{*{20}{c}}
{{{\bm P}_o}}&{\bm \Theta} 
\end{array}} \right]^T}
\end{equation}

The lengths and angles of the actuators are denoted by
\begin{equation}
    {\bm q} = {\left[ {\begin{array}{*{20}{c}}
{{{\bm q}_1}}&{{{\bm q}_2}}&{{{\bm q}_3}}
\end{array}} \right]^T}
\end{equation}
where
\begin{equation}
\begin{array}{l}
{{\bm q}_1} = \left[ {\begin{array}{*{20}{c}}
{l_1^1}&{l_1^2}&{l_2^1}&{l_2^2}&{l_3^1}&{l_3^2}
\end{array}} \right]\\
{{\bm q}_2} = \left[ {\begin{array}{*{20}{c}}
{{d_1}}&{{d_2}}&{{d_3}}
\end{array}} \right]\\
{{\bm q}_3} = \left[ {\begin{array}{*{20}{c}}
{{\eta _1}}&{{\eta _2}}&{{\eta _3}}
\end{array}} \right]
\end{array}
\end{equation}

Therefore, Eq. (\ref{eq:18}) can be represented as 
\begin{equation}
    {{\bm J}_x}\dot {\bm x} = {{\bm J}_q}\dot {\bm q}
\end{equation}
where ${\bm J}_x$ is the direct Jacobian matrix and ${\bm J}_q$ is the inverse Jacobian matrix. ${\bm J}_x$ and ${\bm J}_q$ can be represented as
\begin{equation}
{{\bm J}_x} = {\left[ {{\bm J}_1^1;{\bm J}_1^2;{\bm J}_2^1;{\bm J}_2^2;{\bm J}_3^1;{\bm J}_3^2} \right]_{6 \times 6}} \label{eq:22}
\end{equation}
and
% \begin{tiny}
\begin{equation}
{{\bm J}_q} = {\left[ {\begin{array}{*{20}{c}}
{{{\bm I}_{6 \times 6}}}&{\begin{array}{*{20}{c}}
{e_1^1}&0&0&{f_1^1}&0&0\\
{e_1^2}&0&0&{f_1^2}&0&0\\
0&{e_2^1}&0&0&{f_2^1}&0\\
0&{e_2^2}&0&0&{f_2^2}&0\\
0&0&{e_3^1}&0&0&{f_3^1}\\
0&0&{e_3^2}&0&0&{f_3^2}
\end{array}}
\end{array}} \right]_{6 \times 12}} \label{eq:23}
\end{equation}
% \end{tiny}

The elements in Eq. (\ref{eq:22}) and Eq. (\ref{eq:23}) can be represented as
\begin{equation}
   \begin{array}{l}
{\bm J}_i^j = {\bm s}_i^{j{\rm T}}\left[ {\begin{array}{*{20}{c}}
{{{\bm I}_{3 \times 3}}}&{ - \tilde {\bm u}_i^j{{\bm R}_2}}
\end{array}} \right]\\
e_i^j = {\bm s}_i^{j{\rm T}}{{\bm h}_i}\\
f_i^j = {\bm s}_i^{j{\rm T}}{\bm t}_i
\end{array}
\end{equation}

\begin{figure*}[t]
  \centering
    \includegraphics[width=0.7\linewidth]{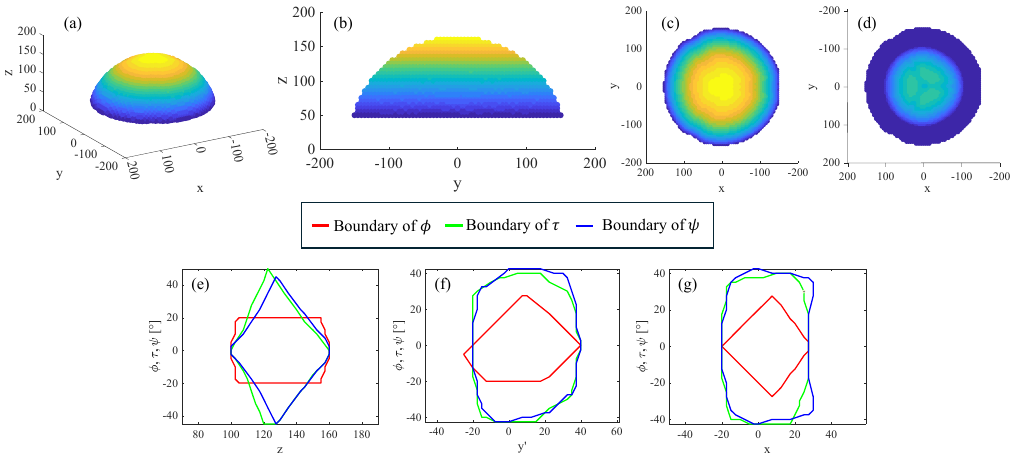}
    \caption{General geometry of $3\text{-}(\underset{\bar{}}{\overarc{\text{P}}}\underset{\bar{}}{\text{P}}(2\text{-}(\text{U}\underset{\bar{}}{\text{P}}\text{S})))$ workspace, when $a=50$, $b=14$, $d_{s}=50$, $r=100$, $l_{\rm min}=114.5$, $l_s=50$ $\eta_{s}=30^\circ$. (a)-(d) Isometric, top, bottom, and side views of position workspace when $\phi=\theta=\psi=0$, (e)-(g) Orientation workspaces.}
    \label{g1}
    \vspace{-3mm}
\end{figure*}

To ensure dimensional homogeneity, the last three columns of the direct Jacobian matrix are scaled by the characteristic length, $L=a/2$. This results in the following form for the dimensionally homogeneous Jacobian matrix ${\bm J}_h$\cite{angeles1992design}. Then, the overall Jacobian matrix can be expressed as
\begin{equation}
    {\bm J} = {\bm J}_h^{ - 1}{{\bm J}_q} \label{eq:26}
\end{equation}

% \vspace{2mm}
\section{Characterization of the Workspace}

It is difficult to directly find a multi-varible function to describe the relationship between the robotic kinematic parameters and its end-effector's Cartesian workspace. However, to tackle the robotic parameter optimization problem in terms of its workspace, we have to find an approach to determine the representation.

In this section, we investigate how the geometric parameters of $3\text{-}(\underset{\bar{}}{\overarc{\text{P}}}\underset{\bar{}}{\text{P}}(2\text{-}(\text{U}\underset{\bar{}}{\text{P}}\text{S})))$ influence its workspace characteristics. The design parameters of $3\text{-}(\underset{\bar{}}{\overarc{\text{P}}}\underset{\bar{}}{\text{P}}(2\text{-}(\text{U}\underset{\bar{}}{\text{P}}\text{S})))$ are listed in Table \ref{tab:Geo}, where $d_{s}$, $l_s$, and $\eta_s$ denote strokes of $d_i$, $l_i^j$, and $\eta_i$ respectively. $l_{\rm min}$ represents the minimum value of $l_i^j$. All lengths are dimensionless, and angles are measured in degrees. 
\begin{table}[!h]
\setlength{\tabcolsep}{1pt} % 只对当前表格生效
	\caption{Design parameters and limits of $3\text{-}(\underset{\bar{}}{\overarc{\text{P}}}\underset{\bar{}}{\text{P}}(2\text{-}(\text{U}\underset{\bar{}}{\text{P}}\text{S})))$}
	\begin{center}
		\label{table_ASME}
		\begin{tabular}{c c c}
        \bottomrule
        Parameter &Description &Value\\
			% put some space after the caption
			\bottomrule
			$a$  &Radius of moving platform &$[10, 100]$\\
			
			$b$ &Distance between $D_1^2$ and $D_1^1$  &$14$\\
			
			$d_{s}$  &Stroke of $\rm \bar P$ on fixed base &$[10, 100]$\\

			$r$ &Radius of fixed base & $100$\\
			
			$l_ {\rm min}$&Min. length of $\rm \bar P$ between $B_1^1$ and $D_1^1$  & $114.5$\\

                $l_s$ & Stroke of $\rm \bar P$ between $B_1^1$ and $D_1^1$& $[10, 100]$\\
			
			$\eta_{s}$ &Stroke of circular $\rm \bar P$ rotation angle &$[10^\circ, 100^\circ]$\\
			\bottomrule
		\end{tabular}
	\end{center}
  \vspace{-2mm}
 \label{tab:Geo}
\end{table}

Based on kinematic analyses, the position workspace and the orientation workspace of $3\text{-}(\underset{\bar{}}{\overarc{\text{P}}}\underset{\bar{}}{\text{P}}(2\text{-}(\text{U}\underset{\bar{}}{\text{P}}\text{S})))$ can be calculated. The position workspace is defined as the set of spatial points that the end-effector can reach when the mechanism’s orientation is fixed. Regarding parallel robots, the volume of the workspace is primarily determined by three critical factors: the maximum extension range of the kinematic chain, the potential mechanical interference between the structural components, and the restrictions of joint motion. For $3\text{-}(\underset{\bar{}}{\overarc{\text{P}}}\underset{\bar{}}{\text{P}}(2\text{-}(\text{U}\underset{\bar{}}{\text{P}}\text{S})))$, the restrictions of joint motion aren't considered as a primary factor affecting the workspace volume and therefore disregard them. 
% \begin{strip}

% \end{strip}

Typically, the orientation workspace is defined as the set of orientations that the end-effector of a mechanism can achieve when its position is fixed. However, this definition only evaluates the orientation capability at a specific point, which limits its ability to provide a comprehensive assessment of the mechanism's overall orientation performance. To address this limitation, a representation method for the orientation workspace based on the symmetry axes of the mechanism is proposed and applied. It is defined as the set of achievable angles $\phi$, $\tau$, and $\psi$ of the end effector along its symmetry axes, $z$, $y'$, and $x$-axes, where $y'$-axis lies in the $xOy$ plane, and the angle between $y'$-axis and $x$-axis is $\theta'$ as shown in Fig. \ref{fig1}. $\tau$ is used to represent the Euler angle of rotation around the $y'$-axis. The new rotation matrix $\bm R'$ is derived in Eq. (\ref{eq:27}) using the Euler-Rodrigues formula \cite{dai2015euler}, where $\bm \theta ' = {\left[ {\begin{array}{*{20}{c}}
{1/2}&{\sqrt 3 /2}&0
\end{array}} \right]^{\rm T}}$.
\begin{figure*}
\begin{equation}
    \bm R' = \left[ {\begin{array}{*{20}{c}}
{{{\theta '}_x}^2 + (1 - s_x^2)c\tau }&{{{\theta '}_x}{{\theta '}_y}(1 - c\tau ) - {{\theta '}_z}s\tau }&{{{\theta '}_x}{{\theta '}_z}(1 - c\tau ) - {{\theta '}_y}s\tau }\\
{{{\theta '}_x}{{\theta '}_y}(1 - c\tau ) + {{\theta '}_z}s\tau }&{{{\theta '}_y}^2 + (1 - s_y^2)c\tau }&{{{\theta '}_y}{{\theta '}_z}(1 - c\tau ) + {{\theta '}_x}s\tau }\\
{{{\theta '}_x}{{\theta '}_z}(1 - c\tau ) - {{\theta '}_y}s\tau }&{{{\theta '}_y}{{\theta '}_z}(1 - c\tau ) + {{\theta '}_x}s\tau }&{{{\theta '}_z}^2 + (1 - s_z^2)c\tau }
\end{array}} \right] 
\label{eq:27}
\end{equation}
\vspace{-5mm}
\end{figure*}
%Generally, we cannot simply assume that a larger range of orientation angles leads to a larger orientation workspace. If the mechanism can only undergo a relatively wide range of orientation angle changes in a narrow direction, its practical application scenarios will be significantly limited. Thus, we 
%
% \begin{figure*}[h]
%   \centering
%     \includegraphics[width=6in]{figs/comparison_new_2.pdf}
%     \caption{(\Rmnum{3})(a-g) Workspaces of $\rm 6 - (\bar P\bar P)U\bar PU$ when $\eta=10^\circ$, (\Rmnum{3})(h-n) Workspaces of $\rm 6 - (\bar P\bar P)U\bar PU$ when $\eta=60^\circ$, (\Rmnum{3})(o) Curve of workspace volume changing concerning $\eta_s$, (\Rmnum{4})(a-g) Workspaces of $\rm 6 - (\bar P\bar P)U\bar PU$ when $d_s=10g$, (\Rmnum{4})(h-n) Workspaces of $\rm 6 - (\bar P\bar P)U\bar PU$ when $d_s=80g$, (\Rmnum{4})(o) Curve of workspace volume changing concerning $d_s$.}\label{fig4}
% \end{figure*}

% what is so-called general geometry? -very unclear.
The geometric workspace representation of the $3\text{-}(\underset{\bar{}}{\overarc{\text{P}}}\underset{\bar{}}{\text{P}}(2\text{-}(\text{U}\underset{\bar{}}{\text{P}}\text{S})))$ in a selected typical configuration case is shown in Fig. \ref{g1}. It should be noted that different colors in (a) to (d) represent the $z$-axis values. In Fig. \ref{g1}(a-c), it can be seen that the main characteristics of the position workspace are a shape resembling an inverted bowl with a top view that appears as a disc. The boundary of the workspace is complete. As shown in Fig. \ref{g1}(d), the workspace has distinct cavities. In addition, its position workspace volume can be calculated as ${\rm V}=3.47\times10^6$. Fig. \ref{g1}(e)-(g) illustrate the orientation capabilities of the $3\text{-}(\underset{\bar{}}{\overarc{\text{P}}}\underset{\bar{}}{\text{P}}(2\text{-}(\text{U}\underset{\bar{}}{\text{P}}\text{S})))$ along the $z$-axis, $y'$-axis, and $x$-axis under the following conditions respectively: $x=0$, $y'=0$; $x=0$, $z=130$; $y'=0$, $z=130$. 

For a better description of orientation capabilities, the torsional capability index (${\rm TI}_1$) and tilting capability index (${\rm TI}_2$) are constructed as follows
\begin{equation}
\begin{array}{l}
{\rm T{I_1}} = \frac{{\Delta {\phi _z}\Delta z + \Delta {\phi _{y'}}\Delta y' + \Delta {\phi _x}\Delta x}}{3}\\
\\
{\rm T{I_2}} = \frac{{\Delta z(\Delta {\tau _z} + \Delta {\psi _z}) + \Delta y'(\Delta {\tau _{y'}} + \Delta {\psi _{y'}}) + \Delta x(\Delta {\tau _x} + \Delta {\psi _x})}}{6}
\end{array}
\end{equation}
where $\Delta \phi_z$ represents the incremental difference in the projection of the boundary of $\phi$ on the horizontal coordinate in Fig. \ref{g1}(e), and $\Delta_Z$ denotes the incremental difference in the projection of the boundary of $\phi$ on the vertical coordinate in the same figure. 
For the geometric representation of the selected configuration in Fig. \ref{g1}, the defined indices can be calculated as: $\rm  TI_1 = 2789.17$ and $\rm TI_2 = 4853.13$.
% It indicates that the tilting capability of the mechanism is superior to its twisting capability.

Based on the proposed analytical method for workspace representation, the effects of changing $a$, $d_s$, $\eta_s$, and $l_s$ on the workspace characteristics of $3\text{-}(\underset{\bar{}}{\overarc{\text{P}}}\underset{\bar{}}{\text{P}}(2\text{-}(\text{U}\underset{\bar{}}{\text{P}}\text{S})))$ are studied while keeping other conditions constant, and several examples are discussed in the following subsections.

\subsection{Influence of $a$}
\begin{figure*}[t]
  \centering
    \includegraphics[width=\linewidth]{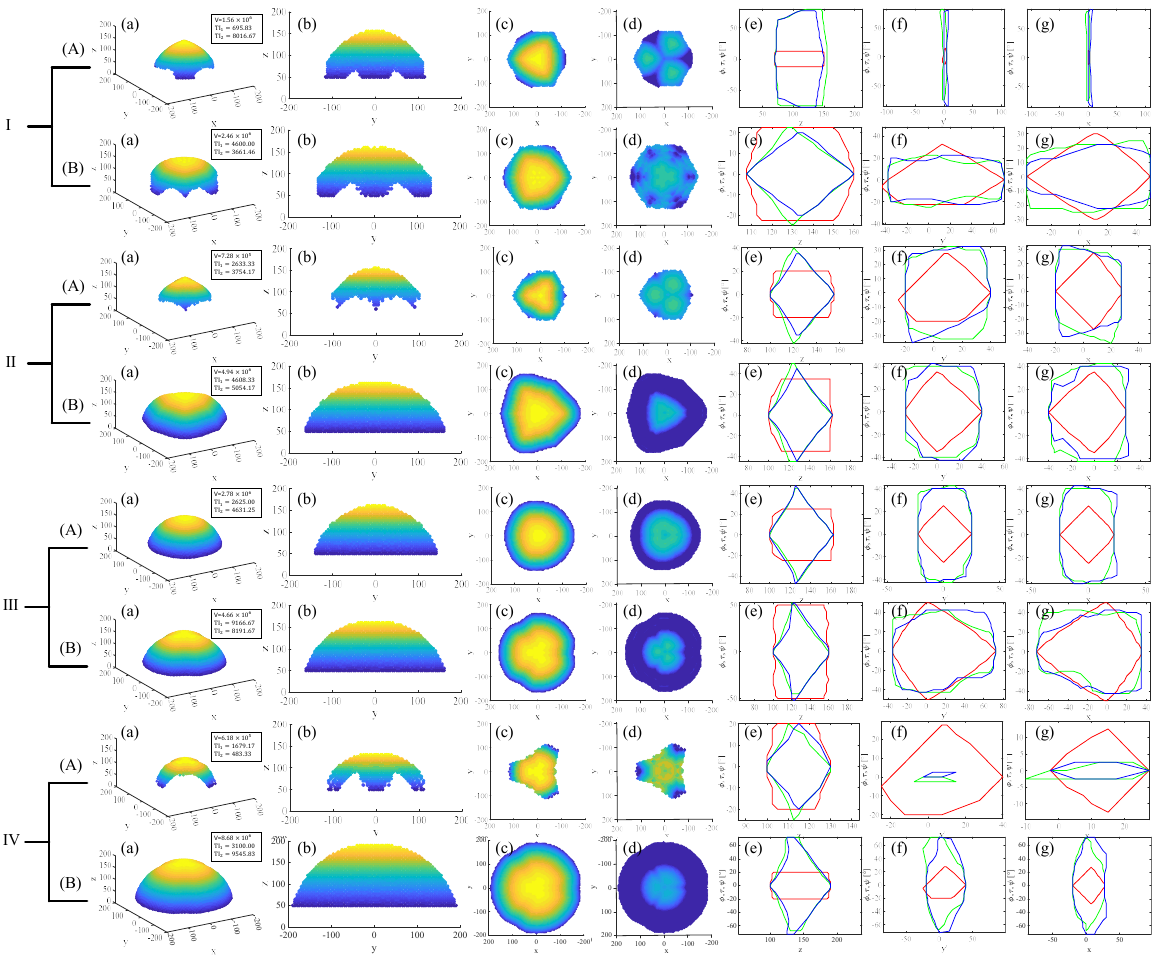}
    \caption{Workspaces of $3\text{-}(\underset{\bar{}}{\overarc{\text{P}}}\underset{\bar{}}{\text{P}}(2\text{-}(\text{U}\underset{\bar{}}{\text{P}}\text{S})))$ when $a=10$, $a=80$, $d_s=10$, $d_s=80$, $\eta_s=10^\circ$, $\eta_S=60^\circ$, $l_s=20$, and $l_s=80$. \Rmnum{1}(A)(a-g) Workspaces when $a=10$, \Rmnum{1}(B)(a-g) Workspaces when $a=80$, \Rmnum{2}(A)(a-g) Workspaces when $d_s=10$, \Rmnum{2}(B)(a-g) Workspaces when $d_s=80$, \Rmnum{3}(A)(a-g) Workspaces when $\eta_s=10^\circ$, \Rmnum{2}(B)(a-g) Workspaces when $\eta_s=60^\circ$, \Rmnum{4}(A)(a-g) Workspaces when $l_s=20$, \Rmnum{4}(B)(a-g) Workspaces when $l_s=80$.}
    \vspace{-3mm}
    \label{fig3}
\end{figure*}

As depicted in Fig. \ref{fig3}\Rmnum{1}, reducing $a$ to $10$ renders the boundary of the position workspace incomplete, significantly decreases the workspace volume, and increases the proportion of internal cavities. In this case, the orientation capabilities of the mechanism are greatly restricted, particularly along the $x$ and $y'$-axes, while its tilting capability about the $z$-axis is enhanced and its twisting capability is weakened. Conversely, when $a=80$, the boundary again becomes incomplete and the volume decreases, with internal cavities occupying a larger proportion of the workspace. However, the orientation capability is more balanced across all three axes, and the torsional capability improves. The curve of workspace volume varying with respect to each parameter is shown in Fig. \ref{fig4}. It can be seen that as $a$ gradually increases, the position workspace volume initially rises and then sharply declines.

\subsection{Influence of $d_s$}
As shown in Fig. \ref{fig3}\Rmnum{2}, When $d_s=10$, the mechanism's position workspace exhibits a significant reduction in volume, an incomplete boundary, a higher proportion of internal cavities, and a diminished horizontal coverage area, while its orientation capabilities remain largely unaffected. When $d_s=80$, the position workspace volume of the mechanism becomes larger, and the external boundary of the workspace remains fully enclosed. Concurrently, the proportion of internal cavities is significantly reduced. In the horizontal plane, the coverage area is expanded, indicating an enhanced reach. Furthermore, the torsional capability of the mechanism increases markedly. Fig. \ref{fig4} shows that as $d_s$ gradually increases, the volume of the position workspace initially grows and then remains almost unchanged.
% \vspace{-10mm}
\subsection{Influence of $\eta_s$}
As shown in Fig. \ref{fig3}\Rmnum{3}, When $\eta_s=10^\circ$, the position workspace volume of the mechanism decreases while maintaining a fully enclosed outer boundary. Additionally, the proportion of internal cavities increases, and the overall orientation capability experiences a slight reduction. When $\eta_s=60^\circ$, the position workspace volume of the mechanism increases, with a fully enclosed boundary and a significantly reduced proportion of internal cavities, resulting in a distinct ``mushroom-like” shape. At the same time, the overall orientation capability is notably enhanced, especially in terms of torsional motion. Fig. \ref{fig4} demonstrates that as $\eta_s$ increases, the volume of the position workspace grows correspondingly.
% \vspace{-3mm}
\subsection{Influence of $l_s$}
When $l_s=20$, the position workspace volume of the mechanism decreases markedly, with a severely incomplete boundary and a significantly increased proportion of internal cavities. Furthermore, the overall orientation capability is considerably weakened, especially in terms of tilting motion. When $l_s=80$, the mechanism’s position workspace volume increases substantially, with expanded coverage in both the horizontal and vertical displacements. The external boundary remains intact, and the proportion of internal cavities decreases significantly. In addition, the overall orientation capability is markedly enhanced, especially in terms of tilting motion. Fig. \ref{fig4} demonstrates that as $l_s$ increases, the volume of the position workspace grows correspondingly.

\subsection{Combined Effect of $a$ and $l_s$ on the Workspace Volume}
\begin{figure}[h]
\vspace{-5mm}
  \centering
    \includegraphics[width=2.7in]{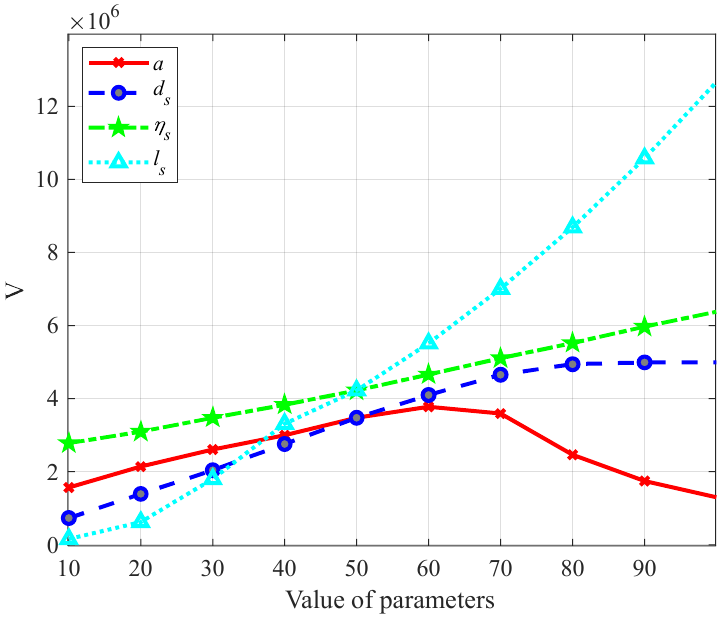}
    \caption{Curve of workspace volume varying with respect to each parameter.}\label{fig4}
\end{figure}

Fig. \ref{fig4} indicates that $l_s$ exerts the greatest influence on the mechanism’s position workspace volume. Because the effects of $\eta_s$ and $l_s$ on the workspace volume increase monotonically,  the combined influence of $a$ and $d_s$ on the position workspace volume of $3\text{-}(\underset{\bar{}}{\overarc{\text{P}}}\underset{\bar{}}{\text{P}}(2\text{-}(\text{U}\underset{\bar{}}{\text{P}}\text{S})))$ is emphasized.

\begin{figure}[!h]
  \centering
  \includegraphics[width=2.6in]{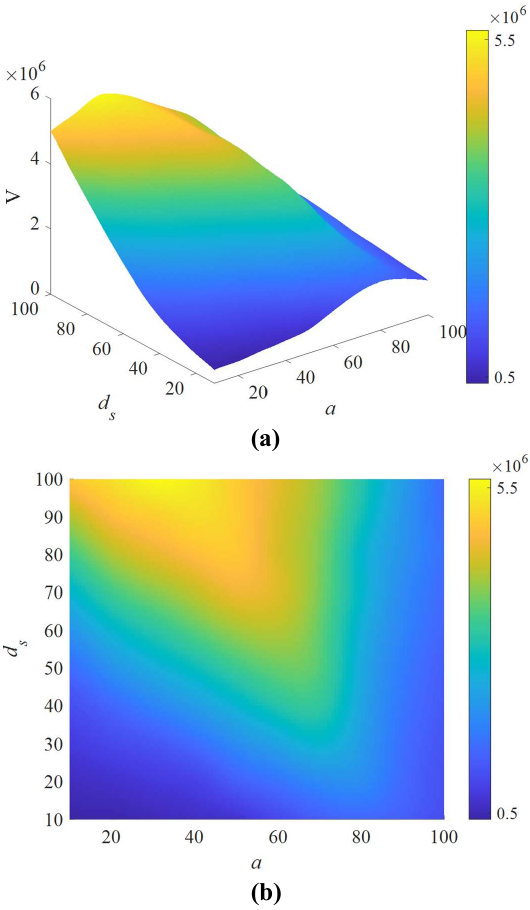}
    \caption{Surface of workspace volume varying with respect to $a$ and $d_s$. (a) Surface that visualizes the relationship between the position workspace volume and the parameters $a$ and $d_s$, (b) Top view of the surface.}\label{fig5}
    \vspace{-5mm}
\end{figure}

Fig. \ref{fig5} shows the variation of the workspace volume ($\rm V$) with respect to the parameters $a$ and $d_s$. It demonstrates the optimal region for workspace volume. This region is characterized by smaller values of $d_s$ and larger values of $a$, as shown by the areas of the surface with the most intense colors, indicating the highest volume values.

\section{Conclusion}
In this paper, a novel redundant parallel mechanism $3\text{-}(\underset{\bar{}}{\overarc{\text{P}}}\underset{\bar{}}{\text{P}}(2\text{-}(\text{U}\underset{\bar{}}{\text{P}}\text{S})))$ is presented, which can be used in various scenarios. To investigate the elementary parameter optimization issue of this redundant parallel mechanism, 
% is very important and challenging thereafter. 
the relationship between key geometric parameters and workspace characteristics is studied. 

The kinematics of $3\text{-}(\underset{\bar{}}{\overarc{\text{P}}}\underset{\bar{}}{\text{P}}(2\text{-}(\text{U}\underset{\bar{}}{\text{P}}\text{S})))$ is firstly established, and its workspace is calculated. To better illustrate, an orientation workspace representation is defined based on the symmetry axes to provide a comprehensive assessment of the mechanism's orientation workspace, and $\rm T_1$ and $\rm T_2$ are constructed to better evaluate the torsional and tilting capabilities of the mechanism.

Subsequently, the effects of $a$, $d_s$, $\eta_s$, and $l_s$ on the mechanism's workspace characteristics are investigated separately. The simulation results indicate that the influence of $a$ and $d_s$ on the position workspace volume exhibits a non-monotonic trend. As $a$ increases, the workspace volume initially expands but eventually diminishes. Moreover, an excessively large or small $a$ will result in an incomplete workspace boundary. In terms of orientation capability, a smaller $a$ weakens the torsional ability of the mechanism while enhancing its tilting ability. When $d_s$ increases, the position workspace volume first grows and then remains nearly constant. An excessively small $d_s$ will result in an incomplete workspace boundary as well. Additionally, a larger $d_s$ enhances the overall orientation capabilities, particularly improving the torsional performance.

Whereas, $\eta_s$ and $l_s$ positively influence the workspace volume, with larger values leading to more extensive workspaces. $\eta_s$ does not affect the completeness of the position workspace boundary but significantly alters its shape. Among all investigated parameters, $\eta_s$ has the greatest impact on the torsional capability of the mechanism. $l_s$ plays a critical role in vertical displacement and tilting ability, with larger values of $l_s$ significantly enhancing the mechanism's vertical movement range and tilting capability, however, an excessive $l_s$ leads to a reduction in torsional performance.

Finally, the combined effects of $a$ and $d_s$ are further explored, revealing an optimal parameter range for maximizing workspace intuitively. In the future, the ways to enhance robotic manipulation capabilities can be investigated, as well as optimal design and performance improvements of $3\text{-}(\underset{\bar{}}{\overarc{\text{P}}}\underset{\bar{}}{\text{P}}(2\text{-}(\text{U}\underset{\bar{}}{\text{P}}\text{S})))$ in specific application scenarios.

\bibliographystyle{IEEEtran}
\normalem
\balance
\bibliography{IEEEabrv, References}

%\end{CJK}
\end{document}